# A New Action Recognition Framework for Video Highlights Summarization in Sporting Events

Cheng Yan[a], Xin Li[b*], and Guoqiang Li[a]

[a]Department of Mechanical & Industrial Engineering, Louisiana State University, Baton Rouge, LA70803; [b]Division of Electrical and Computer Engineering, Louisiana State University, Baton Rouge, LA 70803

*Corresponding author. Tel.:001-225-578-0289; E-mail: xinli@cct.lsu.edu

**Abstract**

To date, machine learning for human action recognition in video has been widely implemented in sports activities. Although some studies have been successful in the past, precision is still the most significant concern. In this study, we present a high-accuracy framework to automatically clip the sports video stream by using a three-level prediction algorithm based on two classical open-source structures, i.e., YOLO-v3 and OpenPose. It is found that by using a modest amount of sports video training data, our methodology can perform sports activity highlights clipping accurately. Comparing with the previous systems, our methodology shows some advantages in accuracy. This study may serve as a new clipping system to extend the potential applications of the video summarization in sports field, as well as facilitates the development of match analysis system.

## 1. Introduction

With the booming of sports industry, broadcast time is increasing every year, E.g., on a worldwide level, the broadcasting time for table tennis surged almost double from 2018 to 2019 (from 3,145 hours to 6,911 hours) (ITTF.com, 2019). It means that coaches and sports analysts possess more video resources than before. However, in order to capture the useful information, they have to repeatedly watch the videos but watching whole game usually are extremely time-consuming. To solve this problem, a video highlights system is needed. In this work, by highlights we mean the video clip sets that have match-irrelevant frames removed (see a more formal definition in Eq.(2)). Currently, for these sports videos, the editing is mainly conducted manually. Human editors have to spend lots of time to trim raw videos, which is time-consuming and labor-intensive. Therefore, it is highly desirable to have an effective video trimming and annotation system to automate this process.

So far, annotation of videos involving human actions have been used for multiple fields, such as games, documentaries, sport videos, art, etc.(Bhaumik et al., 2020; Duchenne et al., 2009; Gammulle et al., 2017; Han et al., 2017; Ji et al., 2013; Johansson, 1973; Laptev & Pérez, 2007; Li et al., 2016; Liu et al., 2009; Shinde et al., 2018; Shotton et al., 2011; Simonyan & Andrew, 2014; Taylor et al., 2012; Wang et al., 2016; Xiong et al., 2020; Zhang et al., 2013; Zhao et al., 2020). Existing approaches can mainly be classified into two types. (1) One is annotation systems with prior knowledge. In these systems, humans are treated as general objects composed of pixels information that can be extracted. For these types systems, a classical method is two-stream convolutional neural network (CNN) (Ji et al., 2013; Shinde et al., 2018; Simonyan&Andrew, 2014; Wang et al., 2016; Xiong et al., 2020; Zhao et al., 2020). To be specific, the space information can be easily extracted from RGB frames and then input as a stream in CNN network. Similarly, for the temporal information, researchers often use another CNN network to collect necessary information. A popular way is the optical flow method (Simonyan&Andrew, 2014), which captures the motion of the object by calculating the motion of image intensities in an image set. Besides that, the temporal information can also be gathered by Long short-term memory method (LSTM) (Gammulle et al., 2017; Li et al., 2016), which makes predictions based on time series data. For the annotation systems with prior knowledge, pixel sets supply enough information and hence this annotation system are easy to be classified. However, these systems do not consider the interior affiliations for these pixels and which leads to an interior defect. (2) The second category includes systems with prior knowledge or skeleton tracking-based annotation systems (Han et al., 2017; Johansson, 1973; Shotton et al., 2011; Taylor et al., 2012; Zhang et al., 2013). In these systems, the prior knowledge is adopted, i.e., the main bodies in the videos by default are human beings. By leveraging the prior human skeleton information, these systems can reasonably obtain the affiliation in the information. In other words, by connecting the dozens of key points of human's skeleton, human behaviors can be effectively tracked. As a result, the training processes are relatively fast because only a small amount of information is extracted and then works as input. However, one important disadvantage is that the skeleton often cannot distinguish the target human and other humans.

To date, for the very popular sports, such as football, basketball, baseball, etc., some generic machine learning annotation systems have been developed (Meng et al., 2017; Piergiovanni et al., 2018; Tsunoda et al., 2017). Nevertheless, for the racquet turn-based sporting events, the generic methods have their own inherent limitation. That is, there are more events in racquet sports activities, thus which bring much classification errors as well as affect the extension of the automate annotation systems. Despite the previous successes in this domain, accuracy is still in dire need of solution in the field for sports video segmentation. In the previous works, the accuracy is often below 90% (Chakraborty et al., 2016; C. Liu et al., 2009; Tang et al., 2011). In order to achieve higher accuracy, significant manual work is still needed. To deal with mis-detected rallies, human editor has to browse the whole video to locate the missed one to further make up the clipped video. Therefore, improving the clipping accuracy in this domain is highly desirable.

In this study, our aim is to develop an efficient sports highlights editing system. We propose to utilize the three-level prediction model based on two open source structures to automatically recognize players' action, and then efficiently summarize the sports highlights. To this end, we address a three-level prediction algorithm in Section 2, and then show experiments that compare our approaches and previous studies in Section 3. Finally, some conclusions are drawn in Section 4.

## 2. Method

### 2.1. Problem Formulation and Terminologies

As the name implies, a *highlights* summarization is aggregate that composed of all rally segments (where break (non-playing) segments is removed). As known, given a video sequence $S$

$$S = \{S_{p_1}, S_{np_1}, S_{p_2}, S_{np_2}, S_{p_3}, S_{np_3}, S_{p_4}, S_{np_4}, S_{p_5} ...\} \quad (1)$$

where $S_p$ and $S_{np}$ represent the rally (playing) and break (non-playing) segments in the video respectively (see Fig.1). Its highlight generation reduces to generating a partitioning of $S_p$

$$S_P = \{S_{p_1}, S_{p_2}, S_{p_3}, S_{p_4}, S_{p_5} ...\} \quad (2).$$

For example, in racquet games, one highlight interval begins with a service action and ends with scoring. A series of match-irrelevant actions such as picking up the ball and walking back to the field/table, and so on, are break segments, and can be removed from the highlights.

In order to obtain the video sequence $S_p$, four procedures are executed.

(1). Low-level prediction: Boolean decision for every player (see details in 2.2).

(2). Middle-level prediction: Boolean decision for every frame (see details in 2.3).

(3). Because of the inevitable prediction errors in procedure (1), some noises are made, i.e., some of type I errors and type II prediction errors appear, which require us to conduct the high-level prediction. Therefore, we continue making Boolean decision for every short video segment based on (2) (see details in section 2.4).

(4). To save time for clipping procedure, an action sequence merge is then executed to obtain the final high clip (see details in 2.5).

The structure schematic diagram for a highlight generation process of a typical sports video is shown in Fig. 1 and the detailed processes are addressed in section 2.2-2.5.

## 2.2. Player Action Prediction

In this study, we choose table tennis as the target sports. As presented above, existing human action prediction approaches can be divided into models with and without prior knowledge. To date, the most popular method for these two types are YOLO (Redmon et al., 2016, 2018, 2017; Shinde et al., 2018) and OpenPose (Cao et al., 2017; Niklaus et al., 2017; Piergiovanni et al., 2018) respectively. To be specific, YOLO v3 is an open source structure for real time object detection, which was firstly developed by Redmon et al. in 2016 (Redmon et al., 2016). The basic structure is shown in Fig. 2(a). Its final output is a tensor with dimension $N \times N \times (5 \times B + C)$, where $N \times N$, B, C are numbers of grid cell of image, number of bounding boxes and numbers of categories, respectively. OpenPose is a person pose detection method developed by Cao et al. (2017). The basic structures are shown in Fig.2(b). Its final output is a human feature vector with dimension up to 135.

We recognize player actions using both YOLO v3 and OpenPose, respectively, and compare their performance. These two systems have some identical and some different components. Both YOLO and OpenPose models use the same manually labeled labels (i.e., playing and non-playing), where playing actions include serves, pushes, loops, etc., and non-playing actions include ball-picking, shoes lacing-up, rest, game preparation, etc. The difference is that YOLO v3 aims to collect pixels information while OpenPose only collects limited key points information as input. Hence, the OpenPose model needs more training data to achieve a comparatively satisfied classification. Specifically, 689 images are manually annotated for YOLO v3 model while 15,216 actions (generated from a 10 minutes match video) are used to train OpenPose model. The machine specification is listed in experiment section (Table 4). In the training, the initial prediction accuracy is tested by predicting 518 images from a segment of a table tennis full match (640x352@30fps for 156s), the details comparison for the initial prediction comparison are listed in Table 1. It is worth noting that the running time for the two models are totally different. The running time for OpenPose (1870s) is about 1.39 times longer than YOLO v3 models (782s). It is because OpenPose needs not only the key point's position, but also interior affiliation of the key points. We noted that both models approximately achieve 80% prediction accuracy, and thus further robustness reinforcement is necessary in the next procedure. We summarize the major advantages and disadvantages of these two methodologies in Table 2.

Table 1 Initial image test results for players based on YOLOv3 and OpenPose models

|  | YOLOv3 | OpenPose |
|---|---|---|
| Training data | Manual annotation for 438 images from the match frames. (718 players tagged: 513 "playing" labels vs 205 "Non-playing" labels) | 15216 training action vectors generated from 10-minutes consecutive video (7948 "playing" labels vs 7268 "non-playing" labels) |
| Test accuracy | For 518 images TN: 76.11% FN:23.89% TP:65.06% | For 518 images TP: 88.11% FP: 11.89% TP:64.78% |

|  | NP:34.94% | NP:35.22% |
|---|---|---|

(TN: true negative; FN: false negative; TP: true positive; NP: negative positive)

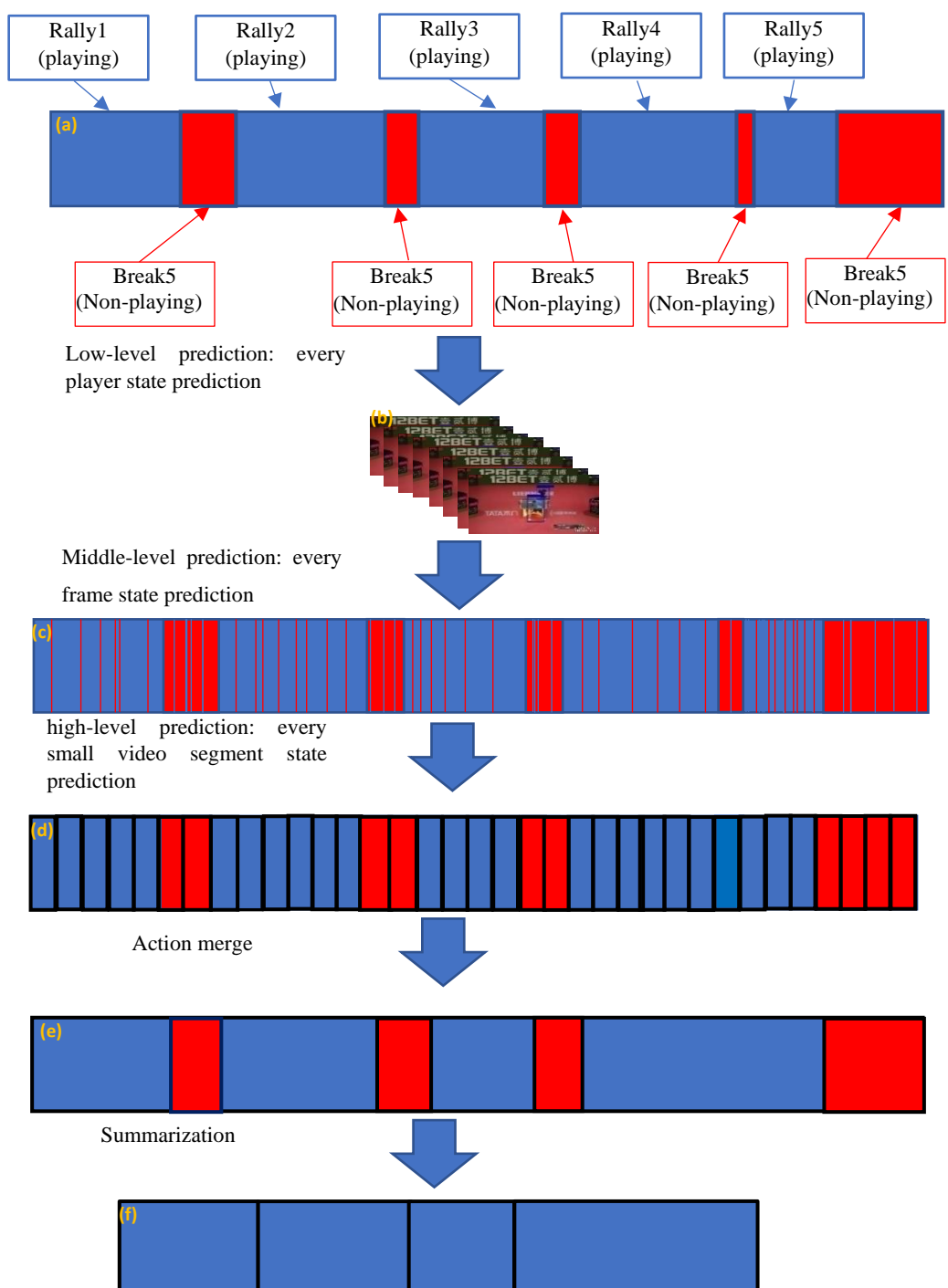

Fig. 1 Pipeline for a highlight generation process of a typical sports video: (a) raw video. (b) low-level prediction set (players) (c) middle-level prediction set (frames) (d) high-level prediction set (short video) (e) video with merged actions (f) final clipped highlight.

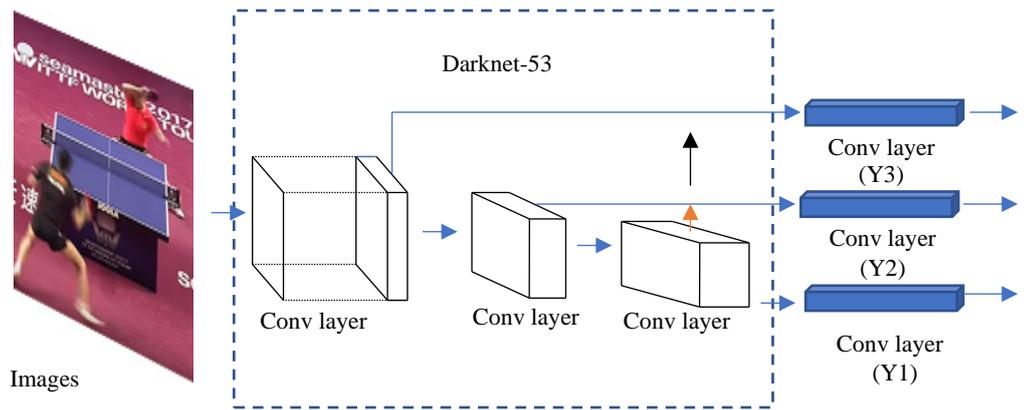

(a)

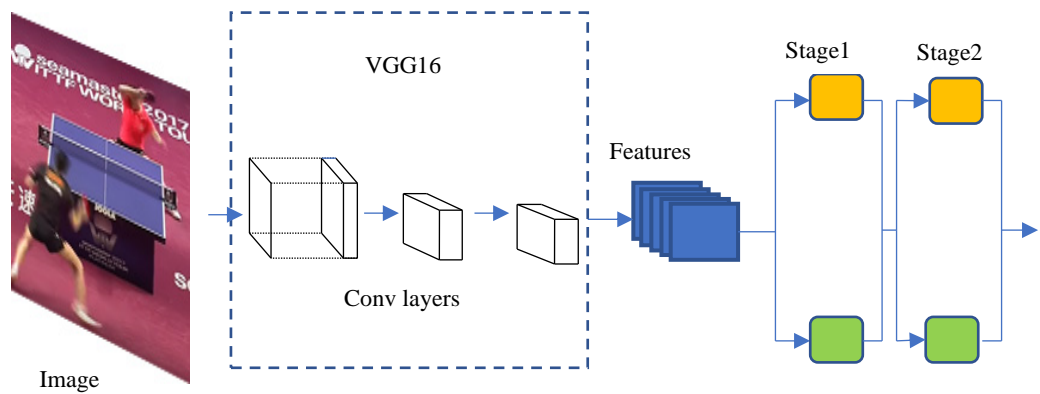

(b)

Fig.2 Basic pipeline structures of two widely adopted video annotation strategies, specifically, (a) YOLO v3, and (b) OpenPose.

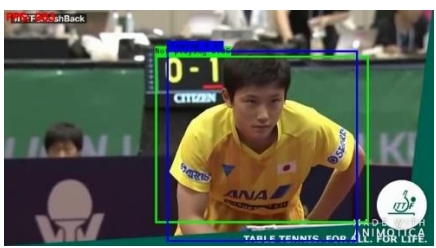           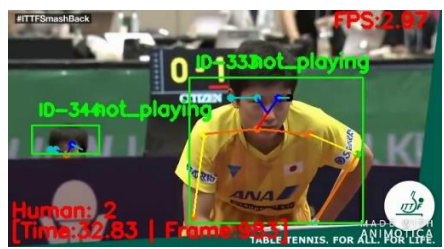

(a)                                                          (b)

Fig.3 Different predictions for a player action from (a) YOLO v3 (two results are obtained for one detected player, where a blue box indicates a "playing" prediction and a green box indicates a "non-playing" prediction, and (b) OpenPose (where the tracked skeleton is visualized and whether the player is "playing" or not is also printed).

Table 2 Comparison for player prediction of YOLOv3 and OpenPose models

|  | YOLOv3 | OpenPose |
| --- | --- | --- |
| Advantage | ● Needs fewer sample annotations.<br>● Has no human number limitation for detection.<br>● Predictions only aim to target human. | ● Training data is easy to generate.<br>● Training process is much fast.<br>● Only generate one prediction for one person. |
| Disadvantage | ● Training is slow.<br>● Could mistakenly taking empire as a player.<br>● Sometimes predictions include both states (labels) for a same player (see Fig 3(a)). | ● Cannot distinguish match-irrelevant players and match-relevant players (see Fig 3(b)).<br>● Cannot identify closed view for athletes.<br>● Errors could occur when too many people in the images |

## 2.3 Single-frame Action Prediction

As indicated above, YOLO v3 and OpenPose use different recognition criteria, and could result in different action predictions for a same given frame. For example, as shown in Fig.3(a), the player's action is mistakenly judged as in both "playing" and "non-playing" states by YOLO v3. The reason is because that YOLO v3 randomly search for pixel set, and which results in the distinctive decisions for the different image areas (even the pixel set are slightly different, see the green box and blue box in Fig. 3(a)). Obviously, these two opposite predictions can bring confusion for the judgment of the frame state. To eliminate this noise, the judging mechanism is proposed as

$$f \text{ is a playing frame if } \sum P_{r_p}(p) > \sum P_{r_{np}}(p) \qquad (3)$$

where $\sum P_{r_p}(p)$ and $\sum P_{r_{np}}(p)$ are probability sum for all "playing" state and "non-playing" state, respectively.

For Openpose, we implement a different judging mechanism. We noted that the humans action label has been uniquely determined in OpenPose. As shown in Fig.3(b), only one "playing" state is predicted by OpenPose instead of multiple states, thus we can directly utilize the Boolean values for "playing state" to predict. Another key thing is that there are totally 2 or 4 players in table tennis match, thus the frame is judged as a "playing" state as long as a "playing" state is found among the first 2 human actions. The judging mechanism is presented as

$$f \text{ is a playing frame if } \exists s_{playing} \text{ for the first 2 humans} \qquad (4)$$

where $s_{playing}$ represents the playing state.

## 2.4 short video classification

In the last section, we perform single frame prediction. Based on this, the Boolean state for short video can be further classified. Theoretically, although the highlight can be directly obtained by clipping all the "playing" frame. However, as presented above, type I error and type II errors are unavoidable, which could result in some key frames missing from a complete shot. In other word, these mistakenly predicted frames would be deleted in the final clipped video and thus confuse audiences. In order to reduce the possibility of wrong decision, we exploit a **voting** scheme. Specifically, the voting scheme can be presented as:

$$s \text{ is a short playing video segment if } \Pr_p(s) > \Pr_c \qquad (5)$$

where $s = [I_1, I_2, ..., I_k]$ is a video sequence composed of $k$ consecutive frames $I_1$ to $I_k$. $\Pr_p$ and $\Pr_c$ represent the confidence levels for the players in a "playing" state and a threshold value. $\Pr_p$ can be simply calculated as

$$\Pr_p = \frac{\sum_{i=i}^{k} B_p}{k} \tag{6}$$

$B_p$ is the Boolean value for the prediction in procedure 2 (section 2.3). By doing that, we obtain an initial time sequence space for short "playing" sequence

$$\{t_{1_{start}}, t_{1_{end}}, t_{2_{start}}, t_{2_{end}}, ... t_{i_{start}}, t_{i_{end}}, ... t_{n_{start}}, t_{n_{end}}\} \tag{7}$$

where $t_{i_{start}}$ and $t_{i_{end}}$ are the start moment and end moment of $i$th short video.

## 2.5 Action merging

From section 2.2 to 2.4, the three-level prediction is conducted. In this section, we conduct an action merging to reduce the clipping time. According to the initial time sequence Eq.(7), we can clip the full match video. However, if we take a closer look at Fig. 1(d), one shot could include one to a few of short videos, which further increases the computational burden. Therefore, we need merge the actions by using a refine process. The elimination algorithm is described as:

$$\text{delete a video segment } (t_{end_i}, t_{begin_{i+1}}) \text{ if } t_{begin_i} - t_{end_{i+1}} < \Delta t \tag{8}$$

where $\Delta t$ is a given constant time interval parameter. Finally, we obtain the second time sequence

$$\{t_{1_{begin}}, t_{2_{end}}, ... t_{i_{begin}}, t_{i+1_{end}}, ... t_{n_{begin}}, t_{n_{end}}\} \tag{9}$$

Apparently, this time sequence could be much shorter than the initial time sequence Eq (6). By using this time sequence, the highlight can be well summarized.

## 3. Experiment and comparison

In this section, 10 table tennis videos with resolution 640×352@25fps or 640×352@25fps are experimented, respectively. Their duration are 6 minutes 41 seconds ~ 59 minutes 11 seconds (video3-video10 are full matches). Following Zhang et al. (2005), three evaluation criterions, i.e., precision P, recall R and combined metric C are presented as following:

$$P = \frac{R_{cd}}{R_d} \quad (10)$$

$$R = \frac{R_{cd}}{R_a} \quad (11)$$

$$C = \frac{2 \times P \times R}{P+R} \quad (12)$$

where $R_{cd}$, $R_d$ and $R_a$ represent the number of correctly detected rallies, detected rallies and actual rallies, respectively. By referring the score caption, we can obtain the number of actual rallies; by counting the final captured number of rallies, numbers of detected rallies and correctly detected rallies can be obtained. It is worth noting that the slow-motion replay is not taken into the statistics. The computer specification and results for 10 videos are presented in Table 3 and Table 4, respectively. As listed in Table 4, the precision P, recall R and combined metric C for YOLO v3 are 95%-100%, 88%-100%, and 91-100% respectively, and the average precision P, recall R and combined metric C are 96.4% and 96.2% and 96.2% respectively; the range of precision P, recall R and combined metric C for OpenPose are 93%-100%, 64%-100% and 76-100% respectively, and the average precision P, recall R and combined metric C are 95.7%, 87.3% and 90.7% respectively. Apparently, comparing with the accuracy in the first procedure (about 80%), the three-level prediction gains a significant improvement.

Table 3 Computer specification and compiling environment

|  |  | specification |
|---|---|---|
| Hardware | Processor | Intel i7-4710HQ |
|  | GPU | GTX 970m 2GB |
|  | Memory | 16GB DDR3 |
|  | Storage | 1TB SSD |

| Compiling environment | System | Windows10 |
|---|---|---|
| | Parallel computing platform | Python3.6.6 |
| | Machine learning platform | TensorFlow-GPU1.14 |

Table 4 Experiment results for two methodology

| Model | Video name | Duration Before clipping | Duration after clipping | Frame numbers | Number of correctly detected rallies | Number of detected rallies | Number of actual rallies | Performance | | |
|---|---|---|---|---|---|---|---|---|---|---|
| | | | | | | | | P | R | C |
| YOLO v3 | Video 1 | 401 s | 222s | 12,030 | 17 | 17 | 17 | 100% | 100% | 100% |
| OpenPose | | | 234s | | 24 | 25 | | 100% | 100% | 100% |
| YOLO v3 | Video 2 | 930 s | 473s | 27,900 | 39 | 40 | 39 | 98% | 96% | 97% |
| OpenPose | | | 578s | | 39 | 40 | | 98% | 100% | 99% |
| YOLO v3 | Video 3 | 2,860s | 1,606s | 71,500 | 88 | 92 | 91 | 96% | 97% | 96% |
| OpenPose | | | 1,572s | | 82 | 86 | | 95% | 90% | 93% |
| YOLO v3 | Video 4 | 1,910s | 929s | 47,750 | 67 | 70 | 69 | 96% | 97% | 96% |
| OpenPose | | | 870s | | 46 | 49 | | 94% | 67% | 78% |
| YOLO v3 | Video 5 | 2,458s | 1,023s | 73,740 | 91 | 94 | 94 | 97% | 97% | 97% |
| OpenPose | | | 1,238s | | 93 | 97 | | 96% | 99% | 97% |
| YOLO v3 | Video 6 | 2,111s | 1,068s | 52,775 | 63 | 66 | 72 | 95% | 88% | 91% |
| OpenPose | | | 737s | | 64 | 67 | | 96% | 89% | 92% |
| YOLO v3 | Video 7 | 3,551s | 1,400 | 88,775 | 96 | 101 | 104 | 95% | 92% | 94% |
| OpenPose | | | 1,781s | | 70 | 75 | | 93% | 67% | 78% |
| YOLO v3 | Video 8 | 2,168s | 777s | 65,040 | 70 | 73 | 71 | 96% | 99% | 97% |
| OpenPose | | | 948s | | 69 | 72 | | 96% | 97% | 97% |
| YOLO v3 | Video 9 | 3,354s | 1,847s | 83,850 | 70 | 73 | 73 | 96% | 96% | 96% |
| OpenPose | | | 1,171s | | 73 | 76 | | 96% | 100% | 98% |

| | | | | | | | | | |
|---|---|---|---|---|---|---|---|---|---|
| YOLO v3 | Video 10 | 3,020s | 1,644s | 75,500 | 113 | 118 | 116 | 96% | 97% | 97% |
| OpenPose | | | 1,435s | | 74 | 79 | | 94% | 64% | 76% |

It can be shown from Table 5 that both of our methodologies can achieve the high clipping accuracies especially YOLO v3 based methodology. The small number of missing rallies could be imperfection of the training sets. Next, we compare our methodology with previous systems for highlight summarization. For example, by using the viewer's facial expression and heart, Chakraborty and Zhang (2016) addressed an automatic identification of sports video highlights on basis of Gaussian mixture models. In the experiment, they used 5 video clips (soccer and tennis) to verify their method, and their average precision, recall and combined metric precision can achieve about 58.58%, 49.58% and 52.67% respectively. By adopting the video-based classification, Gygli (2018) devised a fast video boundary cluster method. In this study, a large training data (with 1,000,000 frames and shot transitions) is used, the precision and recall for the 10 videos (in publicly available RAI dataset) with about 25, 000 frames can achieve averagely 86% and 90.8% for recall and precision, respectively. In addition, by combining the color histogram method (for video) and supervised audio cluster method (for audio) and then a temporal voting scheme, Liu et al. (2009) perform a comparatively improved classification for sports video. For 4 videos (tennis and table tennis) with duration from about 9 - 35 minutes, their systems achieve average 86% and 84.8% for precision and recall. Tang et al. used a supervised learning based on color histogram method, which enable their error rate for cricket video clipping as low as 12.1% (Tang et al., 2011).

It can be observed from Table 5 the accuracy indexes above, we found our framework still has some potential advantage over the previous approaches. First, as indicated by Zhang et al. (1993), the color histogram method is insensitive to the spatial change. However, as known, multiple cameras switching is more common in sports broadcasts and hence plenty of spatial changes is unavoidable. Instead, because our training set has contained player images under multiply camera shooting, multiply camera switching cannot affect the action recognitions in our systems, which means that our methodology can better copy with this scenario. Second, for the histogram method based segmentation methods, a key thing is that the some key frames in this stream must be extracted to obtain the key features for a frame set (Liu et al., 2009; Zhang et al., 2005). Nevertheless, the single frame could not be representative and thus brings a few mistakes. On the contrary, we perform a three-level prediction to compensate the possible error caused by single frame prediction and thus has certain superiority. Third, the color histogram can be affected by the object movement (Zhang et al., 1993) and flash (Gygli, 2018), but which is insensitive to our methodology for both YOLO and OpenPose. Fourth, as indicated by Liu and Dai (2016), the audience excitement can also work as an assistive system to improve the precision. However, this system cannot work for the no-audience games. Therefore, it bears a inherit limitation. Finally, comparing with the previous supervised learning system, our method does not need large training dataset ( in Gygli's study (2018), the training set includes 1,000,000 frames). Overall, our methodologies perform better in video highlights summarization.

## 4. Conclusion

In this study, we have developed two systems for automatically summarizing sports video highlights. These systems will help automate sports video editing and highlights generation. Some conclusions can be drawn:

1. We suggested a high-accuracy sports video highlight summarization framework based on YOLO v3 and OpenPose. Between them, YOLO-based approach performs better than the OpenPose-based one.
2. Our proposed three-level prediction algorithm has significantly improved the summarization accuracy than simply applying frame-by-frame YOLO and OpenPose recognition and frame removal.
3. Although only small amount of training data is used in training our systems, but they result in satisfactory auto-clipping systems in this study.

**Limitations.** We noted that the voting scheme is not always a reliable strategy for the OpenPose-based system. We will explore more effective scheme for it in the future. We will also explore effective sports video summarization based on weakly-supervised or unsupervised learning.